\pdfoutput=1
\documentclass{article}

\usepackage[preprint]{neurips_2024}

\usepackage{fancyhdr}
\usepackage{nth}
\usepackage[utf8]{inputenc} 
\usepackage[T1]{fontenc}    
\usepackage{hyperref}       
\usepackage{url}            
\usepackage{booktabs}       
\usepackage{amsfonts}       
\usepackage{nicefrac}       
\usepackage{microtype}      
\usepackage{xcolor}         
\usepackage{graphicx}
\usepackage{hyperref}
\usepackage{amsmath}
\usepackage{cleveref}
\usepackage{color, colortbl}
\usepackage{tabularray}
\usepackage{multirow}
\usepackage{nicematrix}
\usepackage{caption}
\usepackage{subcaption}
\usepackage{booktabs} 
\usepackage{wrapfig}

\definecolor{mygray}{gray}{0.9}

\title{Pose2Trajectory: Using Transformers on Body Pose to Predict Tennis Player's Trajectory}

\definecolor{mydarkblue}{rgb}{0,0.08,1}
\definecolor{mydarkgreen}{rgb}{0.02,0.6,0.02}
\definecolor{darkred}{rgb}{0.8,0.02,0.02}
\definecolor{darkorange}{rgb}{0.40,0.2,0.02}
\definecolor{darkpurple}{RGB}{111,0,255}
\definecolor{myred}{rgb}{1.0,0.0,0.0}
\definecolor{mygold}{rgb}{0.75,0.6,0.12}
\definecolor{mydarkgray}{rgb}{0.66, 0.66, 0.66}

\author{%
   Ali K. AlShami, Terrance Boult, Jugal Kalita \\
   Computer Science Department, University of Colorado, Colorado Springs \\   \texttt{aalshami@uccs.edu},\texttt{tboult@uccs.edu}, \texttt{jkalita@uccs.edu} \\   \url{https://github.com/alshami52/Pose2Trajectory.git}
}

\begin{document}

\maketitle

\begin{abstract}
Tracking the trajectory of tennis players can help camera operators in production. Predicting future movement enables cameras to automatically track and predict a player's future trajectory without human intervention. Predicting future human movement in the context of complex physical tasks is also intellectually satisfying. Swift advancements in sports analytics and the wide availability of videos for tennis have inspired us to propose a novel method called Pose2Trajectory, which predicts a tennis player's future trajectory as a sequence derived from their body joints' data and ball position. Demonstrating impressive accuracy, our approach capitalizes on body joint information to provide a comprehensive understanding of the human body's geometry and motion, thereby enhancing the prediction of the player's trajectory. We use encoder-decoder Transformer architecture trained on the joints and trajectory information of the players with ball positions. The predicted sequence can provide information to help close-up cameras to keep tracking the tennis player, following centroid coordinates. We generate a high-quality dataset from multiple videos to assist tennis player movement prediction using object detection and human pose estimation methods. It contains bounding boxes and joint information for tennis players and ball positions in singles tennis games. Our method shows promising results in predicting the tennis player's movement trajectory with different sequence prediction lengths using the joints and trajectory information with the ball position.
\end{abstract}

\section{Introduction}
\label{sec:introduction}

Artificial Intelligence (AI) has become indispensable in many domains, including autonomous systems, surveillance systems, and sports analysis \cite{al2022generating}. The power of AI technology has profoundly impacted and transformed sectors such as transportation, surveillance, logistics, manufacturing, and healthcare \cite{siciliano2008springer}. In surveillance systems, developers leverage AI to enhance performance with facial recognition, action recognition, and prediction technologies. Sports analysis involves using data points to evaluate and improve athletic performance by generating insights that can be used to optimize player training and game strategies, including recommending beneficial and optimal ball trajectory and speed. Sports analysis has become increasingly popular due to the availability of wearable devices, videos, and other advanced technologies that make data collection and analysis more accessible.

In sports, AI can potentially revolutionize how athletes and coaches approach performance analysis. A trained AI can efficiently process vast amounts of data in real time. This data-driven approach can help with injury prevention and enable coaches and trainers to make informed decisions, optimize training regimes, identify areas for improvement, and provide invaluable insights into player and team performance.

\begin{figure}[t]
  \centering
  \includegraphics[width=.7\textwidth]{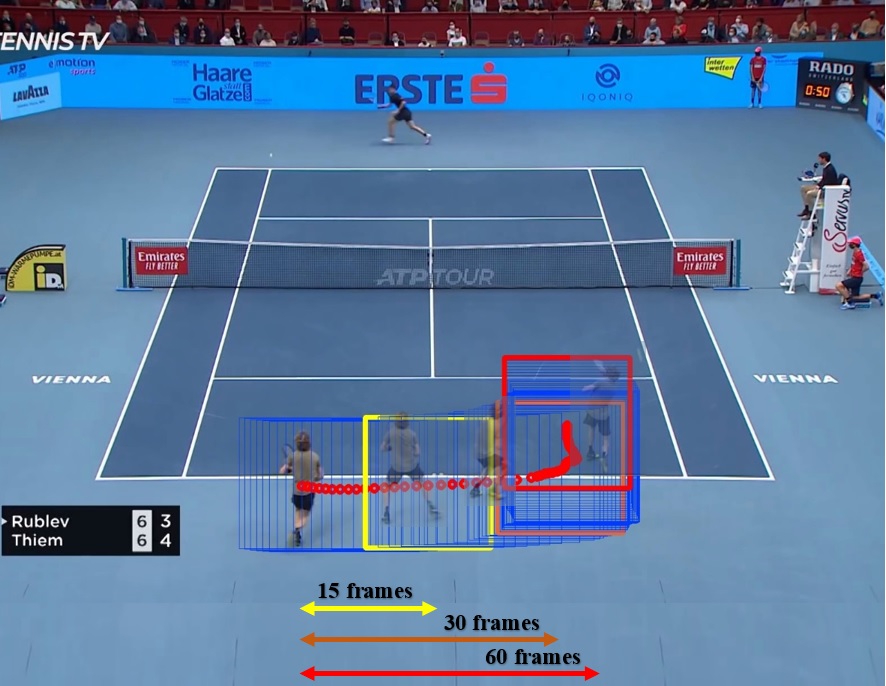}
  \caption{The figure shows the prediction of the future movement trajectory of a tennis player fifteen (250 ms), thirty (500 ms), and sixty (1s) frames ahead (until reaching the ball). The trajectory is predicted as the second player's centroid around which we show a $224 \times 224$ bounding box in future frames. The player appears lighter in future frames. The original image has been taken from videos on the \href{https://www.youtube.com/@tennistv}{TennisTV} YouTube channel.}
  \label{fig:onecol}
\end{figure}

Tennis is a popular sport that first originated in England in the \nth{19} century. In recent years, tennis has seen numerous innovations that help improve referees' decision-making and enhance player performance, including the Hawk-Eye line calling system \cite{owens2003hawk} and telemetry sensors \cite{bergeron2006voluntary}. Processing and analyzing large amounts of data in tennis utilizing AI algorithms can help recognize and predict player actions and movements, forecast future location, speed, and spin of shots, and recognize racquet movements (swing). All this can help trainers evaluate player performance, focus on strengths and weaknesses, and potentially help coaches make more informed decisions during tennis matches. 

Machine Learning algorithms can also help enhance the production of professional tennis game videos by predicting the players' future movement trajectories. The process of recording a tennis match is complex, requiring multiple cameras, skilled camera operators, and advanced broadcasting equipment. Typically, six to eight cameras are strategically positioned around the court, capturing the action from different angles \cite{sheets2011kinematics}. The equipment may include long-range cameras, which provide a comprehensive view of the court and player movements, and close-up cameras, which offer more detailed shots of the players following the ball. All cameras are controlled by skilled operators who follow the action and adjust the angles and zoom levels to capture the best shots. By using the predicted future movement trajectories of the tennis players, these cameras have the potential to automatically track the movement of the players, reducing the need for human intervention and enhancing the accuracy and speed of the recording process.

Predicting the future trajectory of the player in sports is a challenging task. Researchers have attempted to solve this problem in football and basketball \cite{lindstrom2020predicting}\cite{hauri2021multi}. In tennis, people use the players' trajectory information to evaluate performance, style, strategy, and the nature of the movement of the players \cite{pingali2001visualization} \cite{giles2020machine}. 

Enhancing the accuracy of future trajectory prediction for tennis players can enable cameras to autonomously estimate a player's upcoming movements without human intervention. To achieve this, it is crucial to supply predicted data for the player's trajectory at least half a second in advance, taking into consideration the physical constraints of the close-up camera. Typically, cameras used in tennis matches can process multiple requests per second, with a maximum capacity of ten requests per second. However, factors such as processing speed, image resolution, compression settings, and network conditions can impact the camera's response time. Consequently, it is essential to provide predictions at least half a second ahead, or ideally, a full second, to accommodate the camera's response time limitations.

While many research studies on predicting future trajectories for single or multiple individuals rely on the centroid of the bounding box to estimate their movement path, this approach may not be adequate for accurately predicting a tennis player's trajectory. Tennis players exhibit a wide range of movements to reach the ball, which can introduce noise to the centroid-based analysis. To improve prediction accuracy, it is essential to consider specific body parts, such as the arms' joints (determining which one holds the racket), legs' joints, and head, as well as their orientation. This prompts the question: how crucial are the various body parts of a tennis player in making predictions? Examining these elements can offer a more holistic understanding of a player's geometry and motion, ultimately leading to more precise trajectory predictions.

Kalman filter and machine learning methods can be potentially used in sports to track players. However, these algorithms are not explicitly built or reliable for predicting future trajectories over long periods. In our novel approach, we predict the future movement trajectory as a sequence based on the players' body joints, trajectory, and ball position information. In Figure 1, we show the suitability of our method by predicting 15, 30, and 60 frames ahead. We use the encoder-decoder Transformer model in this approach. The encoder part is fed all the joint positions and the centroid points of the players and ball positions in several frames as a sequence, representing the player's past movement. The decoder part predicts the future movement trajectory of the players as a sequence of centroid points. We use a high-quality dataset we created for this work.

In Section 2, we present relevant previous work, including the prediction of future movements in sports. Section 3 describes our prediction system, including tennis player and ball detection, body joint detection, and the use of the Transformer model to predict future movement trajectory. In Section 4, we demonstrate that the body joint information is essential to predict the future trajectory of the players by evaluating alternate models trained on the trajectory information. Finally, Sections 5 and 6 provide information about our dataset and our conclusions.
\section{Related Work}
\label{sec:related_work}
Limited research has been conducted on tennis using computer vision and machine learning to satisfy different requirements. Most of these requirements derive from the needs of referees and tennis players, from beginner to professional, to make better decisions and improve performance, including recognition of the movement of a player's racquet when hitting the ball (swing) \cite{ma2021real}\cite{zhu2007human}, physical simulation of tennis skills from large-scale demonstrations captured in broadcast videos \cite{zhang2023learning}, forecasting future shot locations \cite{wei2016forecasting}\cite{fernando2019memory}, and recognizing tennis players' actions or predicting future actions based on videos or sensor information \cite{ning2021deep} \cite{vinyes2017deep}\cite{polk2019courttime}. Additionally, recent studies have utilized TrackNet-based neural networks \cite{huang2019tracknet} to pinpoint the ball's position, identify court lines, and track player movements for comprehensive analysis of tennis games \cite{rocha2023analysis}.

Analyzing tennis players' trajectories offers valuable insights that can improve predicting their future movements, performance, and strategies. LucentVision, a visualization system, employs real-time video analysis to deliver in-depth information on performance, style, and tactics, all derived from trajectory data \cite{pingali2001visualization}. Another study \cite{giles2020machine} leverages trajectory information to identify player direction shifts. By harnessing these innovative approaches, we can gain a deeper understanding of tennis players' techniques and tactics, ultimately improving predictions and analysis in the sport.

The Kalman filter \cite{kalman1960new}, a widely used estimation algorithm, excels at predicting an object's subsequent movements by incorporating both current measurements and the predicted state from the previous step. Its performance hinges on how closely the model assumptions align with the actual system dynamics and noise properties. When the dynamics are accurately modeled, and the noise is Gaussian, the Kalman filter can precisely estimate the current state. However, its efficacy may decline when predicting multiple steps ahead (e.g., 10, 25, 60 frames) due to various factors, such as model mismatch, accumulated uncertainty, process noise, and nonlinear dynamics. Despite these limitations, the Kalman filter remains a powerful tool for short-term predictions, especially when the underlying assumptions hold true.

In a study by McEwen et al. \cite{mcewenpredictive}, the use of the Kalman filter demonstrated suboptimal performance in predicting the future location of animals based on their centroid positions in images with a resolution of 160x120 pixels. When forecasting five frames ahead, the average error amounted to 80.5 pixels in the x-direction and 19.7 pixels in the y-direction, which is much inferior compared to the prediction of the next frame. Due to the substantial decrease in prediction accuracy as the prediction range increases, we decided against employing the Kalman filter as a method for predicting object positions in this context.

Machine learning methods such as SORT (Simple Online and Real-time Tracking) \cite{bewley2016simple}, DeepSORT (Deep Simple Online and Real-time Tracking) \cite{wojke2017simple}, FairMOT (Fair Multi-Object Tracking) \cite{zhang2021fairmot}, ByteTrack (Multi-Object Tracking by Associating Every Detection Box) \cite{zhang2022bytetrack}, and BoT-SPRT (Bayesian Online Tracking with Spatial Prior-Regularized Trees) \cite{aharon2022bot} primarily focus on tracking objects and maintaining their identities in real-time or near-real-time within video sequences. While these methods are not explicitly designed for predicting long-term future trajectories, such as half a second or a full second ahead, they can offer some degree of short-term prediction. By estimating an object's current state—including its position and velocity—these algorithms can roughly extrapolate the object's trajectory for a few frames into the future. However, their accuracy is likely to diminish drastically for longer-term predictions, as the underlying motion models and assumptions may not remain valid over extended time periods.

Predicting human trajectories as a set of 2D coordinates has been extensively studied in the literature. Early research primarily relied on handcrafted features \cite{mehran2009abnormal}\cite{alahi2014socially}. In contrast, recent studies have used data-driven models such as recurrent neural networks with spatial and temporal features \cite{kosaraju2019social} and Generative Adversarial Networks (GAN) \cite{amirian2019social}\cite{sadeghian2019sophie} with attention mechanisms to predict the future human trajectory of individuals.

Recent studies have used the Transformer to predict future human motion trajectory in various tasks \cite{giuliari2021transformer}. In Li et al. \cite{li2022graph},  the authors used graph-based representations of the pedestrian environment with Transformer and memory replay to improve prediction accuracy. Another approach used non-autoregressive Transformers to simultaneously predict both the trajectory and poses of humans, avoiding the limitations of traditional autoregressive models \cite{mahdavian2022stpotr} \cite{achaji2022pretr}. Group-aware spatial-temporal Transformers, such as \cite{zhou2022ga}, can predict the trajectories of multiple humans by considering group interactions

Finally, Chen et al. \cite{chen2021s2tnet} and Yu et al. \cite{yu2020spatio} used spatial-temporal Transformer networks to predict future trajectories of humans in complex environments. These Transformer-based models have shown promising results in predicting future human movement trajectories, which have potential applications in autonomous driving, robotics, and other fields.

\section{Dataset}
\label{sec:dataset}

Our dataset provides information for tennis players extracted from videos. The information presents 2D keypoints corresponding to the joints of the tennis players and the ball's coordinates in each frame. The videos for the datasets were collected from the \href{https://www.youtube.com/@tennistv}{TennisTV} channel on YouTube for Vienna Open, a professional tennis tournament played on indoor hard courts. Each video presents a point in a tennis match, beginning when one player starts serving until one of the players misses the ball. Tennis players are fast in performing some movements, including serve, front-hand, and back-hand. We convert the videos to images with an average rate of 60 frames per second to cover all the movement information. We modify the shape of our dataset based on the sequence length used for training the model. The multivariate time series dataset has more than one series of observations at each time step. The dataset is available for research.

We have faced some issues creating our dataset. However, we found solutions for the issues. The player detector model aims to detect tennis players, not other people. However, other people are usually involved in tennis videos, including spectators, the referee, and the ball boy/girl. Therefore, searching for instances of class \textit{person} is not a solution. We have tried alternate solutions to avoid capturing other people.
\begin{figure}[t]
  \centering
  \includegraphics[width=0.8\textwidth]{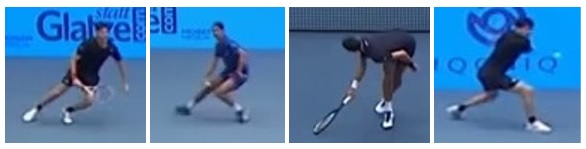}
   \caption{Some of the tennis players' movements that Faster-RCNN could not detect. This is likely because Faster RCNN was pre-trained on non-sports images.}
   \label{fig:onecol}
\end{figure}
The first solution is estimating the background in a sequence of images with median filtering. Median filtering is a non-linear filtering technique that replaces the value of each pixel with the median value of its neighboring pixels within a defined kernel size \cite{dincc2015dt}. However, we found that median filtering needs to be more accurate when other people move. The second solution is to detect the tennis court lines using the Harris corner detector to detect corners and then identify the point nearest each player \cite{derpanis2004harris}. Finally, we used Euclidean distance in each frame to calculate the distance between the \textit{person} object to the closest points.

We encountered issues with the Faster RCNN model's inability to detect certain tennis player movements, as illustrated in Figure 2. This limitation arises from the model's pre-training on the COCO dataset \cite{lin2014microsoft}, which unfortunately lacks sports or tennis-specific images that would have been more suitable for our needs. To mitigate this, we implemented a GUI script to facilitate a review and clean-up process for the image data. This script automatically sorted images with incorrect or poor bounding boxes into a separate file, where they were subsequently manually labeled using AWS SageMaker. As a future direction, we intend to leverage novel image classification and image capturing approaches to better identify unique tennis player movements \cite{shrivastava2023novelty} \cite{yan2021task}.

Another issue that we faced was detecting the tennis ball in videos. Sometimes, the tennis ball disappeared for a few seconds in a part of the video when one of the players hit the ball upward, causing the ball to leave the frame before it came down to the second player, taking a curved trajectory. For example, the ball was invisible for over one hundred frames in one video. We solved the problem using polynomial regression using ten points before and ten after the missing data. After drawing the plot using polynomial regression, we generated points that present the tennis ball positions, taking advantage of the plot.

\section{Approach}
\label{sec:approach}

In this section, we present our system for predicting the future trajectory of a tennis player based on past motion. As shown in Figure 4, our system design incorporates several neural network modules. It consists of four main modules, namely: (1) tennis player detection using Faster RCNN, (2) ball detection using TrackNet, (3) detection of body joints using ViTPose, and (4) future trajectory prediction using Transformer.

\subsection{Detection of the Tennis Players}
Significant works have been published in the last few years to detect objects from videos, including Faster RCNN \cite{girshick2015fast}, YOLO \cite{redmon2016you}, YOLOv6 \cite{li2023yolov6}, SSD \cite{liu2016ssd}, and RetinaNet \cite{lin2017focal}. In this work, we use the Faster RCNN object detection model to detect the tennis players \cite{girshick2015fast}.

Faster R-CNN is a cutting-edge deep learning model for object detection and segmentation, enhancing traditional CNNs with innovative features. It incorporates a Region Proposal Network (RPN), which actively scans an image to produce object-boundary proposals. These proposals are subsequently funneled to a CNN that extracts vital features. A classifier and a bounding box regressor then work in concert to predict the object classes and obtain their precise locations within the image. The synergistic working of RPN and CNN equips Faster R-CNN with accuracy in object detection that markedly surpasses its forerunners. In our methodology, videos from the dataset are transformed into sequences of images, which are then channeled into the Faster RCNN model for latent feature extraction. This model further applies a CNN-based RPN to the unearthed feature maps, generating box proposals. Utilizing the concept of an anchor box for feature maps, RPN emerges as an adept tool for predicting tennis player movements. Each output point materializes a variety of boxes, with each box encapsulating class information and coordinate positions. Specifically, we targeted the class \textit{person} to recognize the tennis players, a process illustrated in Figure 3.

\subsection{Detection of the Ball Position}

Many works have attempted to detect tiny objects such as balls in different sports from video, including TrackNet \cite{huang2019tracknet}, TrackNetV2 \cite{sun2020tracknetv2}, and MonoTrack \cite{liu2022monotrack}. We used the TrackNet model \cite{huang2019tracknet} to detect the tennis ball. 
\begin{figure}[t]
  \centering
  \includegraphics[width=.8\textwidth]{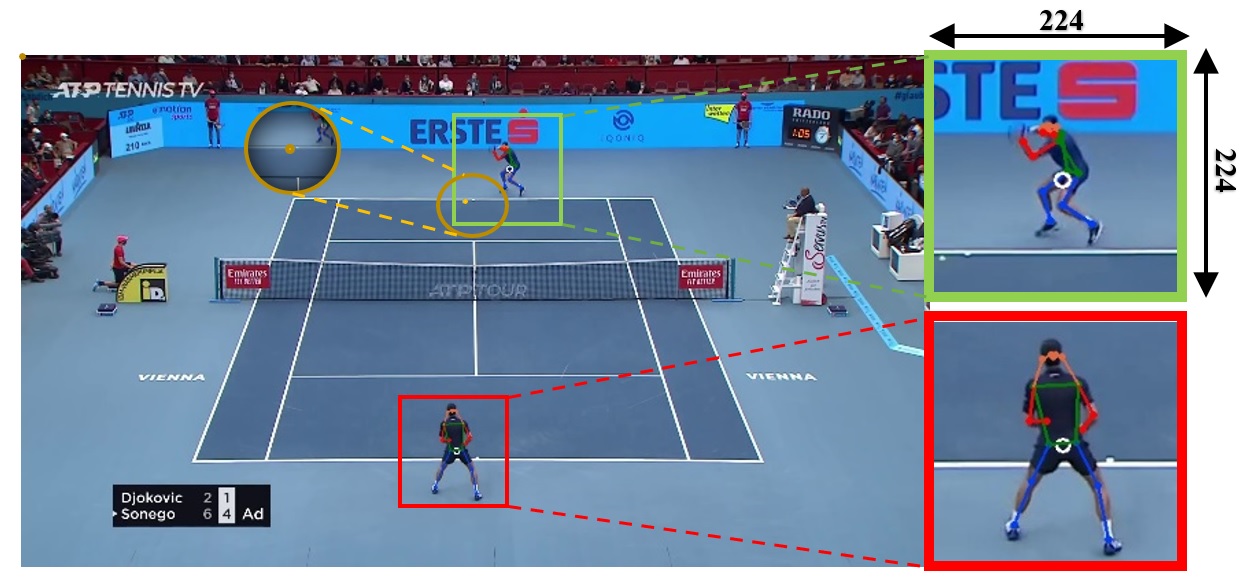}
   \caption{Detecting the tennis player bounding boxes, ball position, and joints using Faster RCNN, ViTPose, and TrackNet. The image has been taken from videos on the \href{https://www.youtube.com/@tennistv}{TennisTV} YouTube channel.}
   \label{fig:onecol}
\end{figure}
Detecting a fast-moving ball in sports videos presents a significant challenge, particularly when the balls appear tiny and can travel at extraordinary speeds. Based on the \href{https://tenniscreative.com/fastest-tennis-serve}{tennicreative} website, the velocity of professional tennis for the first serve often reaches approximately 120 mph for men and around 105 mph for women. To accurately detect a tennis ball hurtling at such formidable speeds, a high-performance model is essential. For our analysis, we employed the TrackNet model to obtain the coordinates of the tennis ball's position within each video frame, a process vividly illustrated in Figure 3. We use the TrackNet model, which is a heatmap-based deep learning architecture meticulously designed to track the elusive movements of tiny tennis balls within video sequences. Distinct from traditional models, TrackNet's training enables it to identify the ball in individual frames while deciphering its flight patterns across successive frames. With the capacity to process images at a resolution of 640 × 360, TrackNet generates a detection heatmap from a single frame or a sequence of frames, ensuring the precise localization of the ball. This capability allows TrackNet to exhibit high precision, even in the realm of public-domain videos.

\subsection{Detection of Body Joints}
To detect the tennis players' body joints, we use a state-of-the-art human pose estimation method called ViTPose \cite{xu2022vitpose}, designed to estimate the 2D and 3D poses of human bodies from 2D RGB images. The model uses a Vision Transformer (ViT)-based \cite{dosovitskiy2020image} architecture for image classification tasks. ViTPose takes an input image and encodes it using a ViT backbone. The ViT backbone is trained to divide the input image into a set of patches, which are then processed by a series of Transformer layers to extract features representative of the entire image. It is trained on large datasets of labeled images, making it a powerful and efficient model for estimating human poses from the 2D image. After detecting the tennis players, we crop the image for each player to size $224 \times 224$ and feed it to the ViTPose model and get the coordinates of seventeen 2D keypoints, $X$ and $Y$, for each keypoint, as output for joint locations such as the elbows, wrists, and knees, as shown in Figure 3.

\subsection{Transformer to Predict Future Trajectory}
The Transformer model has an encoder-decoder deep learning neural network architecture that was first introduced for machine translation \cite{vaswani2017attention}. It has since become the foundational model in natural language processing and computer vision, including the Bidirectional Encoder Representations from Transformers (BERT) \cite{devlin2018bert} and Generative Pre-Training (GPT) \cite{radford2018improving} and the Vision Transformer (ViT) \cite{dosovitskiy2020image}. The model is based on the concept of attention mechanism, introduced by Bahdanau et al. \cite{bahdanau2014neural} to address the bottleneck that arises using a fixed-length encoding vector, where the decoder has limited access to the information provided by the input. The encoder takes an input sequence and generates a hidden representation that captures and learns dynamic contextual interactions in the sequence. The decoder takes the hidden representation of the encoder and generates the output sequence. Since Transformer-based architectures have performed well in sequence-to-sequence tasks, including trajectory prediction, we believe that it is an excellent choice for predicting the future trajectory of tennis players as a sequence from videos. 

\begin{figure*}[t]
  \centering
  \includegraphics[width=1\textwidth]{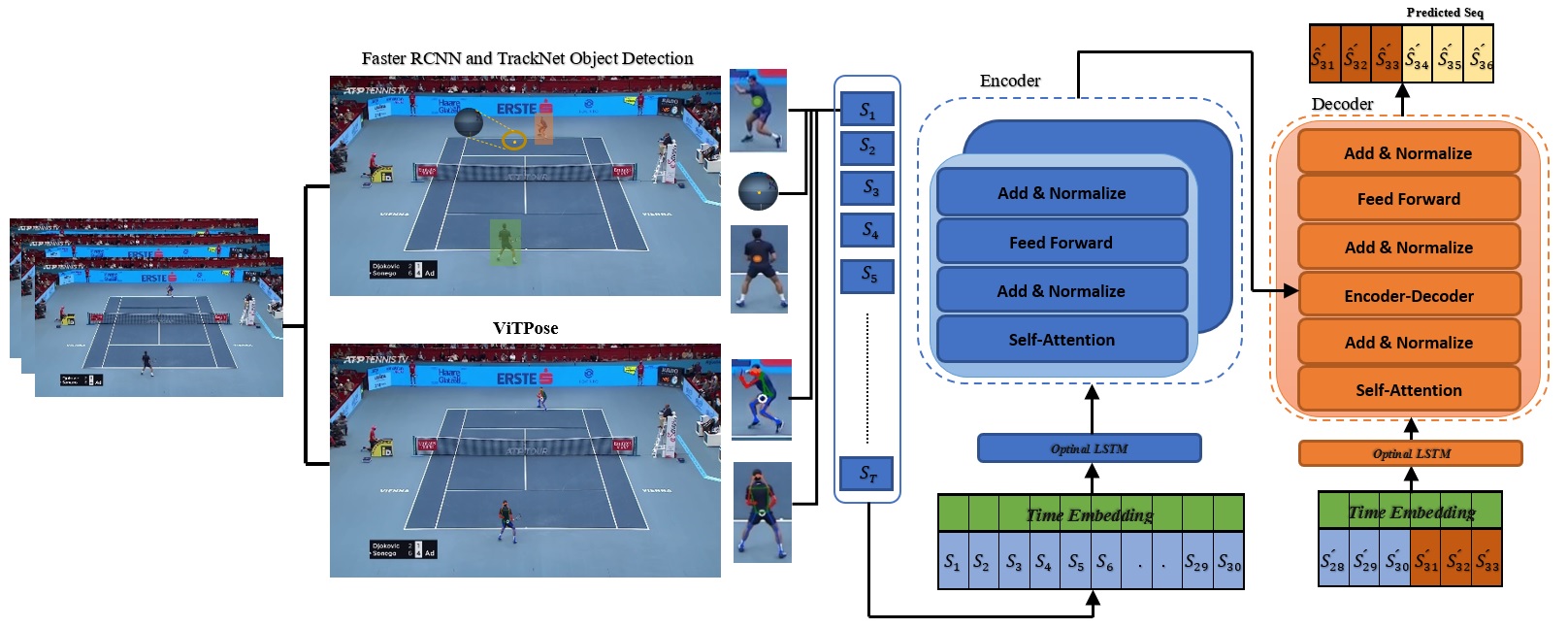}
   \caption{Our encoder-decoder Transformer model to predict a tennis player's trajectory. The encoder encodes players' centroids, joint positions, and ball positions with time information. The decoder takes part of the encoder information along with time information and predicts future centroids of the player at several future time points. $S_i$ represents a feature set of 74 values at time $i$, including player centroids, player joint positions, and ball positions. $S^{'}_i$ represents a set of two values indicating $X$ and $Y$ positions of player centroid that we want to predict at time $i$.}
   \label{fig:onecol}
\end{figure*}
In this subsection, we present our Pose2Trajectory approach, a novel method that predicts the future trajectory of a tennis player as a sequence of 2D coordinates. Our method performs a sequence-to-sequence transformation using an encoder-decoder Transformer architecture. Specifically, the encoder is fed several sets of features in a time sequence, with each set representing information about a past frame that our model considers when making the prediction. These features include the coordinates of joint positions and the trajectory of the tennis players in terms of centroid locations as well as ball positions. 

The decoder also takes several sets of features in time sequence as input. The decoder input consists of two parts: i) selected information (the subset to be predicted) from several frames from the immediate past that were input to the encoder also, and ii) selected information (the subset to be predicted) from a few frames that follow the frames in i). Note that there are two types of information in Figure 4. $S_i$ represents $74$ features of information about frames $i$ containing body joints, ball, and player centroid, whereas ${S}^{'}$ contains only two features corresponding to body centroid. Thus, the sets of features in the decoder input and target represent the centroid locations of the tennis player we want to predict. The approach we use is called the technique of \textit{teacher forcing} to aid the learning process, where the decoder is fed with previous values of the actual output sequence instead of its prediction \cite{williams1989learning}. This is a proven method to help  a sequence-to-sequence model learn from the correct sequence of ground-truth outputs during training, improving accuracy and generalization to new data. By utilizing these input features to describe the players' motions, our model can capture complex relationships between them, resulting in better forecasting results.

As previously mentioned, the Transformer encoder plays a crucial role in processing the input sequence by generating a set of encoded representations that capture the temporal dependencies and patterns in the sequence. In our method, the decoder input incorporates some small subsets of features that the decoder needs to predict from the encoder input, while the decoder target sequence also contains features from the decoder input, resulting in improved encoder decoder processing. This technique has been successfully applied in time series forecasting studies \cite{wu2020deep}.

For example, in Figure 4, we predict a tennis player's movement six frames into the future, given an encoder sequence of 30 frames. In this example, the input to the encoder is a sequence of sets of features, given as $[S_1, S_2, S_3, ..., S_{30}]$, while the decoder input is $[S^{'}_{27}, ..., S^{'}_{33}]$. The goal of the decoder is to output a sequence of sets of features that corresponds to the predicted future trajectory, namely $[\hat S^{'}_{30}, ..., \hat S^{'}_{36}]$. When using Model Family 1 for this example, as mentioned in Table 1, $S$ represents the centroid coordinates of both players, denoted as $ S = C_1, C_2$. On the other hand, with Model Families 2 and 3, $S$ represents the centroid coordinates of the players and their respective joint positions, denoted as $S = C_1,J_1,C_2,J_2$. In Model Family 4, $S$ includes all the information in model families 2 and 3 plus the position of the ball: $S = C_1,J_1,C_2,J_2,B$. Note that $S^{'}$ and $\hat S^{'}$ represent the centroid coordinates of the player we want to predict. For example, if we predict player 1, then $S^{'} = C_1$ and $\hat S^{'} = C_1$, respectively.

Temporal information is necessary to predict future human movement because it allows the system to understand how movement patterns change over time. By analyzing tennis players' historical body joint positions, we can identify patterns and trends that can be used to forecast future movement trajectories. In our work, we use the Time2Vector model \cite{kazemi2019time2vec} to convert the timestamp inputs into a vector representation by mapping the time steps to a high-dimensional space using Fourier transforms\cite{bracewell1986fourier}. The Time2vector model can be presented mathematically as follows:
\begin{equation}
t2v(\tau)[i] =   \begin{cases} 
    \omega_i \tau + \varphi_i \hspace{1.7cm}  \textrm{if }\  i = 0 \\  
    F( \omega_i \tau + \varphi_i), \hspace{0.6cm}  \textrm{if  } \  1\leq i\leq k
    \end{cases} 
\end{equation}
\noindent where $t2v(\tau)[i]$ is the $i^{th}$ element of $t2v(\tau)$, $F$ is a periodic activation function and $\omega_i$ and $\varphi_i$ are learnable parameters.

Using the time vector only in the encoder Transformer is common since the encoder is responsible for processing the input sequence and capturing the relevant information in time. By encoding the time dimension of the input sequence using Time2Vector, the model can capture the temporal patterns in the input data and use this information to generate accurate predictions. Adding the time patterns to the decoder of the sequence-to-sequence model can help the model generate more accurate predictions by providing information about the time at which each prediction should be made. In our case, using the time vector in the decoder part of our model is also helpful. We concatenate the time vector representation with the input sequences before we feed it to the encoder and decoder. We also used an LSTM model with the encoder and decoder to make the prediction smoother. As shown in Figure 5, the prediction results for $X$ and $Y$ are smoother using the LSTM in our architecture,making it better for camera control. LSTMs can make predictions smoother due to their ability to learn and maintain internal states that represent long-term dependencies in the data. In the context of time series prediction or sequence-to-sequence problems, smoother predictions are desirable to reduce noise and capture the overall trend of the data. LSTM networks can achieve this by retaining important information from previous time steps and incorporating it into the current prediction.
\begin{figure*}[t]
  \centering
  \includegraphics[width=.9\textwidth]{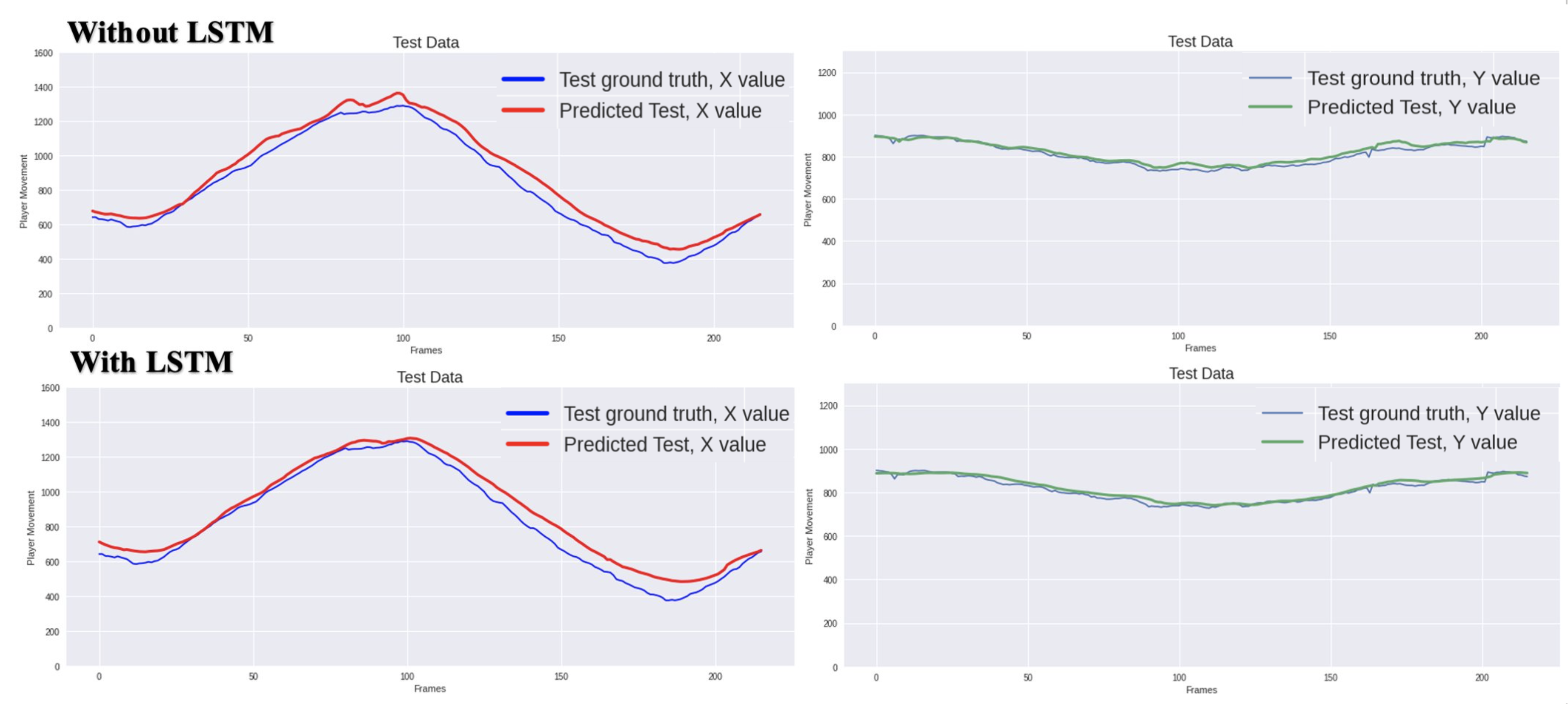}
   \caption{The prediction results for $X$ and $Y$ coordinates with and without using the LSTM in our architecture.}
   \label{fig:onecol}
\end{figure*}
As shown in Figure 4, this work uses two Transformer encoder layers. Each layer contains two sub-layers: a self-attention and a fully connected feed-forward sub-layer, where each sub-layer is followed by a normalization layer. The output of the Transformer encoder feeds to the Transformer decoder to provide the decoder with the necessary context and information to generate accurate predictions for the output sequence. We use a Transformer decoder that contains three sub-layers. In addition to the two sub-layers in each encoder layer, the decoder inserts a third sub-layer to apply self-attention mechanisms over the encoder output. The decoder mask has also been used in this work to introduce a sequence mask that ensures that the model only attends to the past time steps during inference and not to the future time steps that need to be predicted. The mask is necessary to prevent the model from using future information when making predictions, which violates the causality principle in time series forecasting. The Mean Square Error (MSE) loss function has been used with the Adam optimizer \cite{kingma2014adam} with hyperparameters $\beta_1 = 0.9$, $\beta_2 = 0.98$ and $\epsilon = 10^{-9}$. Dropout and weight decay have been used for this work to prevent overfitting with 30 epochs.

\section{Experiments and Results}
\label{sec:experiments}

This section presents the performed experiments and results for different encoder-decoder models. Our models have been trained on sequences of different input lengths to predict a set of 2D coordinates of the centroids representing the future trajectory as a sequence. The input features include the trajectory information of the players, as well as all joints and ball positions. 

Table 1 displays the use of four distinct model Families 1 through 4. Model Family 1 was trained exclusively on tennis players' trajectory information to predict a single player's movement trajectory. Model Family 2 used trajectory information and body joint data for all players to predict the player's trajectory. Model Family 3 employed the same features as Model Family 2 but with the additional use of a mask to account for occlusions. Even though Model Family 2 shows good results when feeding all the body joint features to the encoder, the mask has improved Model Family 3 performance. In Model Family 4, we used trajectory information and body joint data for both players plus the ball position with the mask to predict one player's trajectory.

To evaluate the performance of the Model Families, we calculate the Mean Euclidean Distance Error (MEDE) \cite{danielsson1980euclidean} between the ground truth of the data with the prediction results in pixels using this equation:

\begin{equation}     
MEDE =  \frac{\sum^{T}_{t=1}\sqrt{ {(x_t - \hat x_t)}^2 + 
 {(y_t - \hat y_t)}^2}}{T} .
\end{equation}
\newcommand{\noindentn}{\noindent} The variables $x_t$ and $\hat x_t$ represent the ground truth and prediction, respectively, for the $X$ coordinate. Similarly, the $y_t$ and $\hat y_t$ represent ground truth and prediction for the $Y$ coordinate. The summation is taken over $T$ time points.
\begin{table*}
   \caption{Evaluating the prediction results of the first player with different Model Families that train on sequences of different lengths and numbers of features using the Mean Euclidean Distance Error (MEDE). We calculate the MEDE between the ground truth of the data with the prediction results in pixels.} 
   \label{tab:example}
   \small
   \centering
   \begin{tabular}{|c|c|c|c|c|c|c|c|c|}
   \toprule\toprule
   {} & {} & \multicolumn{7}{|c|}{Predicting Seq-len} \\
   \toprule
   \textbf{Model Name} & \textbf{Training Seq-len} & \textbf{50 ms} & \textbf{100 ms} &\textbf{150 ms} &\textbf{200 ms} &\textbf{250 ms} &\textbf{500 ms}&\textbf{1 s}\\ 
   \midrule
   {} &500 ms & {79}&{95}&{110}&{90}&{91}&{102}&{93}\\
   {Model Family 1} &750 ms&{89}&{154}&{149}&{177}&{131}&{134}&{142}\\
   {} & 1 ms&{105}&{108}&{129}&{132}&{158}&{189}&{124}\\
   \midrule
   {} &500 ms&\textbf{36}&\textbf{50}&\textbf{47}&{71}&{84}&{91}&{109} \\ 
   {Model Family 2}&750 ms&{64}&{71}&{86}&{82}&{99}&{148}&{119}\\
   {} & 1 ms&{85}&{93}&{89}&{103}&{119}&{126}&{147}\\
   \midrule
   {} &500 ms&{46}&{52}&{57}&\textbf{53}&\textbf{64}&{83}&{86} \\ 
   {Model Family 3} &750 ms&{42}&{63}&{82}&{88}&{123}&{93}&{98} \\
   {} & 1 ms&{54}&{59}&{69}&{101}&{99}&{118}&{128} \\   
   \midrule
   {} &500 ms&{58}&{46}&{60}&{66}&{77}&\textbf{48}&\textbf{67}\\ 
   {Model Family 4} &750 ms&{55}&{69}&{71}&{80}&{88}&{71}&{78} \\
   {} & 1 ms&{61}&{73}&{115}&{114}&{85}&{77}&{70}\\ 

   \bottomrule
   \end{tabular}
\end{table*}
We can think of our approach as utilizing close-up cameras to continuously track the tennis player by predicting future movement trajectory with a 224 x 224 boundary box, as shown in Figure 1. We assume that the predicted trajectory remains inside the bounding box up to the error of 48 pixels when predicting 30 frames ahead (500 ms) in Model Family 4. Thus, it keeps cameras capturing the player's body movements. In addition, we evaluate the performance of Model Family 4, which has been trained on 500 ms sequences, in predicting sequences of different lengths (250 ms, 500 ms, 1 s), using the ground truth centroid coordinates X and Y as the reference for comparison, as shown in Figure 6.

\subsection{Analysis}
Our experiments evaluated four model families to predict the future trajectory of a tennis player. Model Family 1 performed the worst, while Model Family 2 showed improvement in short trajectory prediction, predicting 50 ms, 100 ms, and 150 ms ahead, as shown in Table 1 with bold font.

Model Family 2 showed bad predictions with longer sequences. However, using a mask, Model Family 3 significantly improved the prediction of long sequences. Applying the mask in our work helped obtain the best results when predicting 200 ms and 250 ms into the future by preventing information leakage during training. By masking future time steps, the model was forced to learn how to generate accurate predictions based on the information available at each time step, resulting in more robust and generalizable representations.

In the final part of our experiments, we incorporated the position of the tennis ball into the input sequence for Model Family 4. This allowed the model to capture the relationship between the players' movements and the ball's position, leading to more accurate predictions, especially when the ball's position influenced the player's movement. As a result, Model Family 4 showed the best results with less error in pixels using the MEDE for predictions 500 ms and 1 s ahead, with 48 and 67, respectively.

Our findings suggest incorporating contextual information like ball positions and masking techniques can significantly improve models' performance for predicting long sequences of a tennis player's movement trajectory.

\begin{figure}[t]
  \centering
  \includegraphics[width=.5\textwidth]{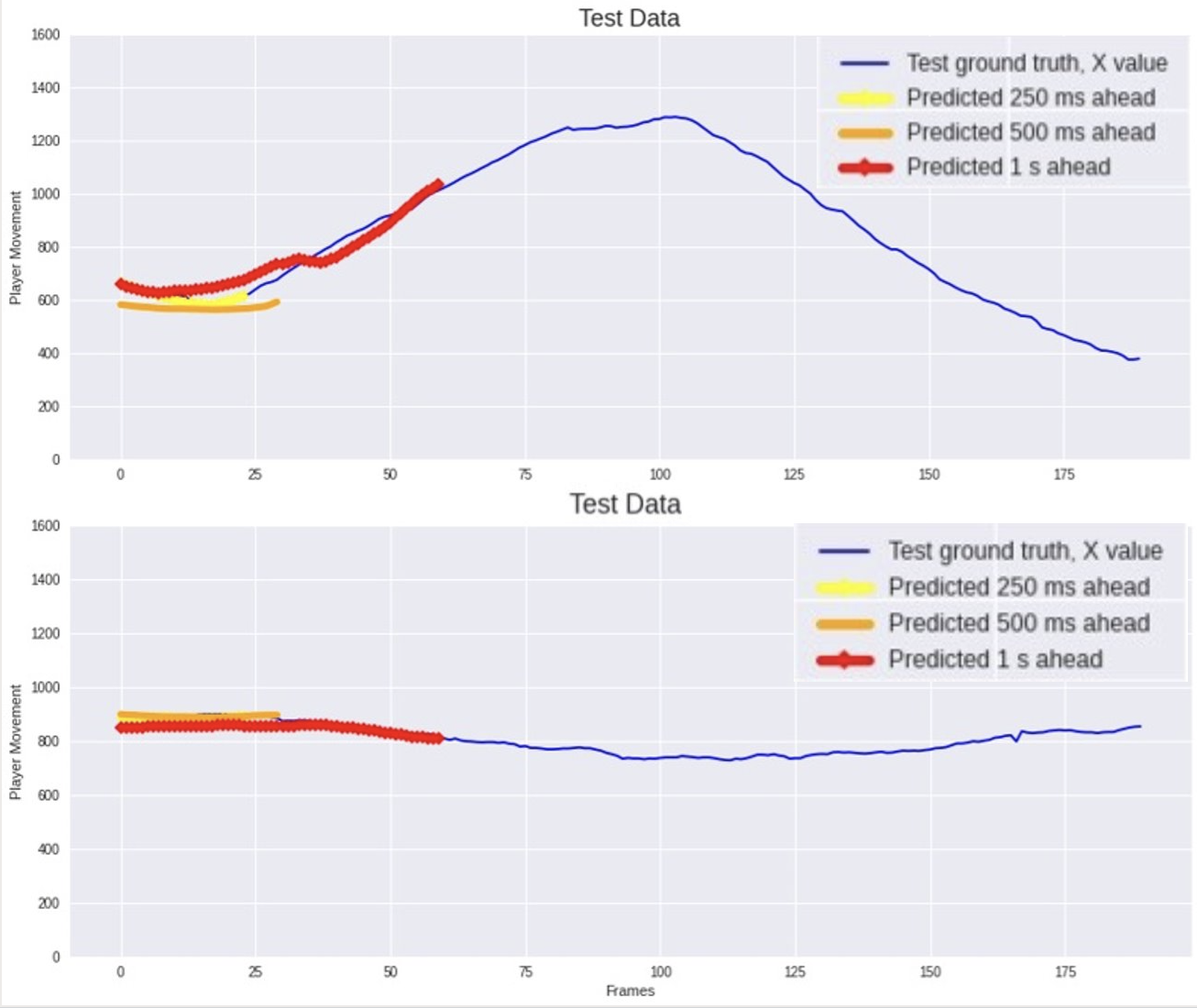}
   \caption{Compare the predicted results of 250 ms, 500 ms, and 1 s ($X$ and $Y$) of model Family 4 with the ground truth of the centroid coordinates of the tennis player.}
   \label{fig:onecol}
\end{figure}
\section{Conclusion}
\label{sec:conclusion}
In this work, we have proposed a novel approach for predicting the future movement trajectory of tennis players by leveraging various sources of information, such as the player's joints, past trajectory, and ball positions. Our approach has the potential to enable cameras to track the tennis players by automatically predicting their centroid location without any human intervention, using an encoder-decoder Transformer model. Our model incorporates a variety of available information, including body joints, trajectory, and ball position, to most accurately predict the player's future trajectory. In particular, The model Family 4 shows the least error for predicting long sequences and uses all the features with a decoder mask.

\small
\nocite{*}
\bibliographystyle{unsrt}
\bibliography{ref}

\end{document}